\documentclass[letterpaper, 10 pt, conference]{ieeeconf}  

\IEEEoverridecommandlockouts                              

\usepackage{multirow}
\usepackage{bbding}
\usepackage{pifont}
\usepackage[nointegrals]{wasysym}
\usepackage{caption}
\usepackage{subcaption}
\usepackage{amsmath}
\usepackage{array}
\usepackage{color}
\usepackage[normalem]{ulem}
\usepackage{hyperref}
\usepackage{nomencl}
\makenomenclature
\usepackage[ruled,vlined]{algorithm2e}
\usepackage{amsmath,amssymb,amsfonts}
\usepackage{algorithmic}
\usepackage{graphicx}
\usepackage{textcomp}
\usepackage{wrapfig}
\usepackage{mdframed,lipsum}
\usepackage[T1]{fontenc} 
\usepackage{booktabs}
\usepackage{mwe}
\usepackage{soul}
\usepackage[table, dvipsnames]{xcolor}
\usepackage{array,booktabs}

\usepackage{cite}
\usepackage{tikz,booktabs,xcolor}

\newcommand{\definemycolor}[2]{%
        \definecolor{#1}{HTML}{#2}%
        \expandafter\def\csname#1html\endcsname{#2\relax} 
        }
\definemycolor{green}{98D992}%
\definemycolor{yellow}{DDE891}%
\definemycolor{white}{FFFFFF}%

\title{\LARGE \bf
Projecting Robot Intentions Through Visual Cues:\\ Static vs. Dynamic Signaling}

\author{Shubham Sonawani, Yifan Zhou and Heni Ben Amor
\thanks{S.~Sonawani, Y.~Zhou, and H.~Ben~Amor are with the School of Computing and Augmented Intelligence, Arizona State University
{\tt\small \{sdsonawa, yzhou298, hbenamor\}@asu.edu}}%
}

\begin{document}


\maketitle
\thispagestyle{empty}
\pagestyle{empty}

\begin{abstract}
Augmented and mixed-reality techniques harbor a great potential for improving human-robot collaboration. Visual signals and cues may be projected to a human partner in order to explicitly communicate robot intentions and goals. However, it is unclear what type of signals support such a process and whether signals can be combined without adding additional cognitive stress to the partner. This paper focuses on identifying the effective types of visual signals and quantify their impact through empirical evaluations. In particular, the study compares static and dynamic visual signals within a collaborative object sorting task and assesses their ability to shape human behavior. Furthermore, an information-theoretic analysis is performed to numerically quantify the degree of information transfer between visual signals and human behavior. The results of a human subject experiment show that there are significant advantages to combining multiple visual signals within a single task, i.e., increased task efficiency and reduced cognitive load.  
\end{abstract}

\section{INTRODUCTION}
Among the many roles future robots are envisioned to assume, one particularly challenging role is that of a human teammate. In such collaborative scenarios, robots have to provide continuous assistance to a human partner, while also ensuring a mutual understanding of the cooperative task and its individual elements. Consequently, there has been significant research interest in generating \emph{interpretable robot behavior} that allows a human partner to better anticipate the goals, intentions, and future actions of a robot for the purpose of fluent teaming. For instance, the Roadmap for U.S. Robotics report highlights that ``humans must be able to read and recognize robot activities in order to interpret the robot's understanding"~\cite{christensen2009roadmap}. 

\begin{figure}[th!]
    \centering
    \includegraphics[width=3.4in]{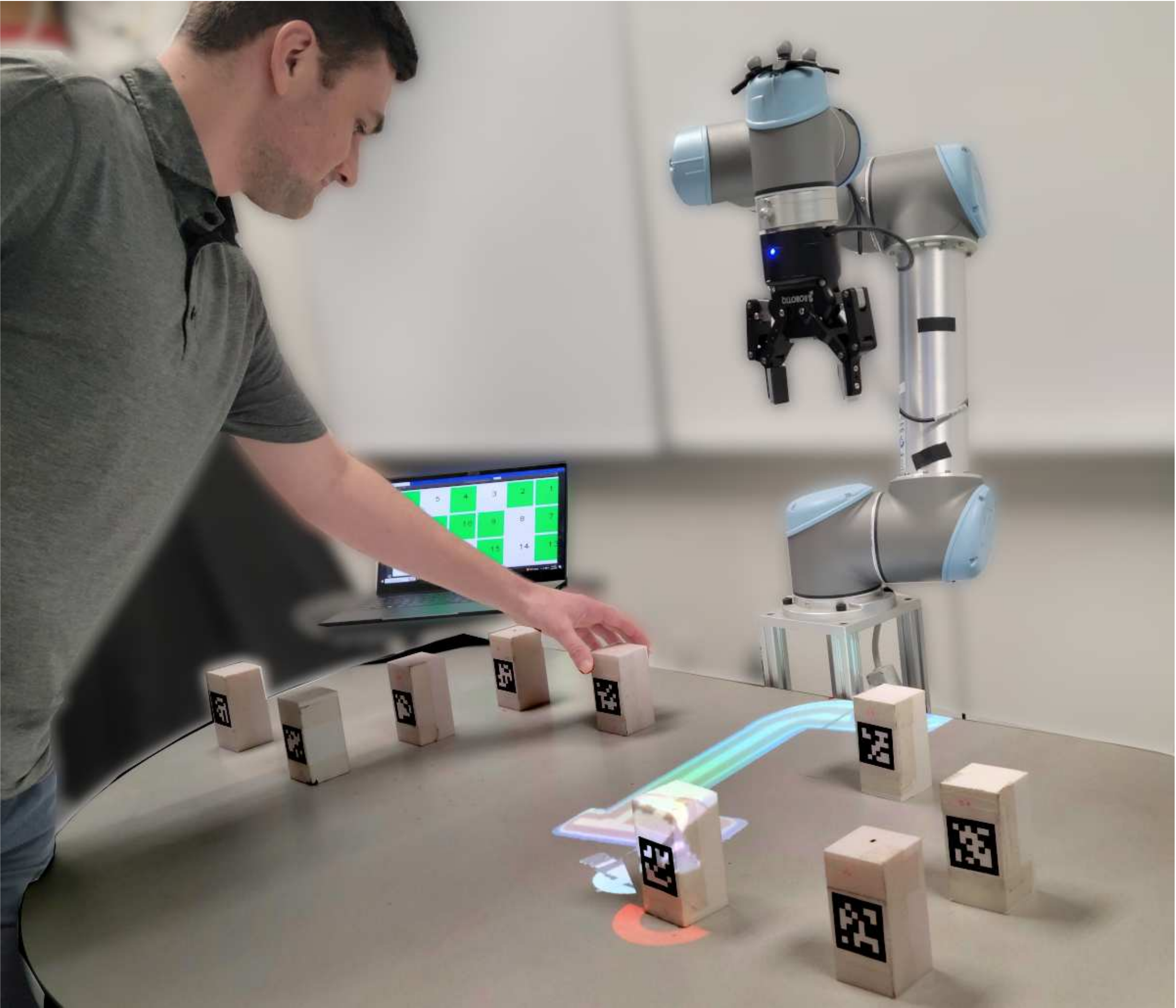}
    \caption{The setup of our experiments: a human subject is tasked with sorting all cubes indicated by green squares, while a robot sorts all the cubes shown by white squares on the laptop screen. Here, a static cue (red semi-circular disc), and a dynamic cue (digital twin of the robot) are projected onto the physical environment showing the current goal of the robot.}
    \label{fig:teaser}
    \vspace{-0.7cm}
\end{figure}

To achieve such a shared mental model, several approaches use \emph{implicit cognitive cues} to communicate robot intentions, e.g., by adjusting robot motion to elicit a specific interpretation from a human observer~\cite{dragan13, zhang2017plan, han2021}. Alternatively, other approaches use \emph{explicit cognitive cues}, e.g., visual, haptic or auditory signals to improve human understanding of robot intentions~\cite{mutlu2016cognitive,fiore2013toward,dumora2012experimental,lackey2011defining}. To this end, the use of augmented and mixed-reality techniques has gained considerable attention in recent years~\cite{burdea2003virtual, schuemie2001research, costa2022augmented,costanza2009mixed, rokhsaritalemi2020review, hughes2005mixed}. The work in~\cite{ramsundar} presented a robot system which projects information about a collaborative task directly into the shared workspace -- a mixed reality approach. For example, by projecting a warning sign onto a particular object in the scene, a robot may identify a goal object it intends to manipulate. As a result, the environment becomes a canvas for the display of perceptual messages that can rapidly be processed by the human visual cortex. Another way to provide visual cues is discussed in \cite{dianatfar2021review, sibirtseva2018comparison, ostanin2020human, williams2018virtual, chadalavada2020bi, liu2017understanding}, e.g., using virtual reality glasses or head-mounted displays which augment the environment around humans.  Various works have provided ample evidence that such visual projection of intent improves critical dimensions of human-robot collaboration tasks, e.g., efficiency, fluency, and trust. 

\begin{figure*}[t!]
    \vspace{0.2cm}
    \centering
    \includegraphics[width=\linewidth]{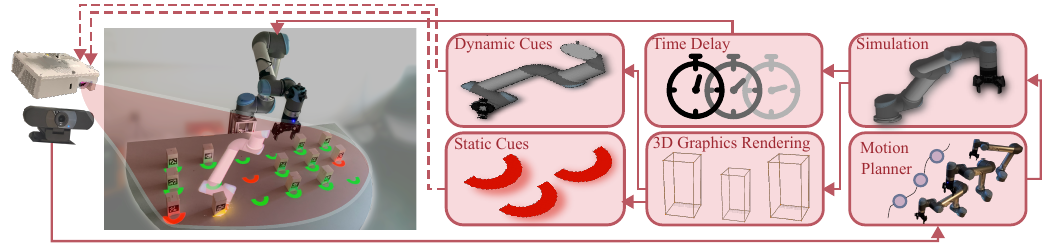}
    \caption{Overview of the system architecture and experiment setup. Information about the current state of the real world environment is captured using a camera. In turn, a robot motion planner generates the intended following actions. The result is used to produce 3D visualizations of either static visual cues or dynamic motion cues. Finally, generated visual signals are projected into the world using a projection device. In this work, we use the projected signals only for the robot actions. However, the designed mixed reality system can also show other signals such as human goals (green) shown in the left image. }
    
    \label{fig:overview}
    \vspace{-0.65cm}
\end{figure*}
However, there are still critical questions and challenges that are not well understood. In particular, it is unclear how to choose and design visual signals so as to achieve the desired transfer of information between the robot and the human partner. To date no objective measures of information transfer have been established that could answer such questions. In a similar vein, it is unclear whether static visual signals (e.g. signs) are preferable to dynamic ones (e.g. and an animation of a moving object or robot). Insights from related fields such as Semiotics~\cite{taniguchi2016symbol,coradeschi2013short} and Human-Computer Interaction (HCI)~\cite{katona2021review,holden2022human} can partially be transferred to this scenario but do not fully address the relationship between human, robot and physical environment.    

This paper extends prior work on intention projection by investigating the effectiveness of different types of signals in human-robot collaboration tasks. Specifically, we compare static and dynamic cues, as well as combinations thereof, to assess their impact on user performance. To this end, we discuss a collaborative sorting task in which a human user has to anticipate the robot's motion to avoid potential collisions or conflicting sub-goals, e.g., reaching for the same object as the robot. We discuss a dynamic signal in which a simulated digital twin of the robot (projected using mixed reality) performs future actions ahead of the real, physical one. Accordingly, the human user can visually anticipate the upcoming motion of the robot. We contrast this mode with a static signaling type, in which the target object is highlighted using a stationary visual cue. Rather than focusing on the actions of the robot, this mode focuses on the underlying goal object only. By comparing the performance of users in dynamic and static conditions, we aim to understand whether different types of signals affect the human user in different ways. The contributions of this paper can be summarized as follows:
\begin{itemize}
    \item A mixed reality system for static and dynamic signaling of robot intention. The system features a novel mode for dynamic signaling that leverages a projected digital twin of the robot to preview upcoming actions.
    \item A human subject experiment focusing on multiple projection modes, along with an extensive analysis of subjective and objective metrics. In addition, information-theoretic approaches are used to numerically quantify the amount of information transferred to the user through visual cues. 
    \item An open-source release of the proposed system along with all necessary components to reproduce the described experiments or investigate other visual cues.  link: \href{https://github.com/ir-lab/IntPro.git}{https://github.com/ir-lab/IntPro.git}
    
\end{itemize}

\section{RELATED WORK}
In human-robot collaboration, efficiently communicating a robot's intentions to a human co-worker is a well-known challenge~\cite{gong2018}. Inherently, humans are excellent at understanding and communicating to each other through nonverbal cues. However, this ability does not apply when the human tries to predict a robot's motion or trajectories in a collaborative setting. To date, robots lack the skill, physical subtlety, and human-like appearance to provide such nonverbal cues effectively. Thus, substantial research has been devoted to different modalities such as gesture, gaze, and haptic feedback to overcome this communication gap~\cite{neto2019gesture,waldherr2000gesture,cabrera2021cohaptics,brosque2021collaborativewelding}. All of these modalities have shown promising results in improving human-robot collaboration. However, humans are visual creatures and can often process explicit visual cues faster than implicit or indirect cues. In addition, visual cues can be co-located with the environment or context they refer to. To convey the presence of a dangerous object, for example, we can display a hazard sign in close proximity or on top of it. Virtual or mixed-reality frameworks provide an excellent technological platform for visualizing such cues in an engaging and interactive fashion. For example, a survey and overview of different modes of visualization that can be used in industrial applications can be found in \cite{costa2022augmented}. The work in  \cite{walker2018communicating} uses a head-mounted display to show augmented reality signals indicating an indoor drone's path and navigation points. In a similar vein, \cite{rosen2019communicating} uses head-mounted displays to depict robot workspace and trajectory information. A technical requirement to achieve this effect is virtual reality headsets or see-through displays. However, as a side effect of this requirement, users may develop fatigue or nausea during the operation of the task. Similarly, such setups may make involving multiple humans in the interaction scenario difficult since one  headset per participant is needed. Alternatively, a mixed-reality setup can be used. Specifically, a camera-projection stereo system can be used to accurately project information on 3D surfaces without needing external hardware such headset. The concept of providing the robot intentions via mixed-reality projections framework called intention projection was previously explored in \cite{andersen2016projecting}. This work compared different interfaces, such as textual descriptions, monitor displays, and projections in human-robot collaboration tasks. Results show that users find the projection interface to be more reliable and effective. Similarly, \cite{ramsundar} discusses how projected patterns can form a rich visual language used in a specific context or domain, e.g., collaborative assembly. Independently of the specific technical implementation, all of the above papers build upon the same principle -- the use of virtual, augmented, or mixed-reality to visually communicate information to a human interaction partner~\cite{williams2018virtual,williams2019mixed,hamilton2021s}. Unlike previous work, the main focus of our paper is on the types of signals and their effectiveness in transferring information between agents rather than on the technical details of intention projection. The aim is to gain insights into various signaling strategies that can be used in any system. By providing a methodology for studying the effectiveness of different types of visual cues, the paper provides a framework for designing and improving intention projection systems. 

\section{Methodology}
In this section, we will first describe the task used throughout the paper. Thereafter, we will provide details regarding our intention projection system and two types of visual signals that are representative of a broader class of signals. Furthermore, details about the transfer entropy and its use for finding a causal relationship between the robot and human is explained.  

\subsection{Collaborative Task}
\label{subsec:task}
In order to investigate the effects of visual signals, an object sorting task was designed, see Fig.~\ref{fig:teaser}. The task involves sorting eighteen 3D-printed cubes placed on a tabletop surface. Human participants are asked to sort twelve cubes while an autonomous robot (Universal Robot UR5) is assigned six cubes. Sorting involves picking assigned cubes, one at a time, and placing them in designated areas. For participants, this designated area is a second table placed on their left-hand side. By contrast, the robot has to place cubes into boxes to its left and right. Visual signals are used throughout the task to visualize the target objects or the motion of the robot. 

\subsection{Intention Projection System}
\label{subsec:intention_projection}
Following the rationale of this paper, visual cues of robot intention are projected into the joint workspace. Fig.~\ref{fig:overview} depicts these cues and provides a detailed overview of our intention projection system. First, a webcam observes the current scene and tracks the location of physical objects on a table. Tracking is performed using simple fiducial markers~\cite{wang2016apriltag}. The resulting scene information (object positions and orientations) is sent to a motion planner (developed using \cite{bruyninckx2001open}) in order to generate valid robot motions to reach the intended next goal. In turn, the plan is simulated, and the resulting data is used to generate visualizations of future states of the system (robot or object). Finally, the generated visual cues are projected into the physical environment using a projection device. For calibration purposes, we use the  method explained in \cite{Moreno2012}: the camera-projector system can be treated as a stereo system consisting of a monocular camera and a projector (inverse camera) by projecting multiple binary patterns on a checkerboard and, in turn, computing intrinsic and extrinsic parameters of the projector in relation to the camera via homographies. To render different projections, we leveraged the combination of the Unity game engine and OpenCV. Furthermore, to provide seamless communication between the robot and the simulation, we used the Robot Operating System (ROS) Noetic version.

\subsection{Visual Signal Types} 
\label{subsec:visual_signal_types}
As mentioned before, we are interested in contrasting two types of signals, i.e., Static and Dynamic visual cues of robot intent. Static visual signals, depicted as red semi-circular disks, highlight the target cube to be picked next by the robot. As Dynamic visual signals, the projection system displays a continuous animation of the intended robot motion. It is important to note that robot motions are shown \emph{ahead of time} -- the user sees a preview of upcoming actions. An example of this Dynamic mode is shown in Fig.~\ref{fig:shadow}. A virtual twin of the physical robot is projected onto the table. This virtual robot moves ahead of time and provides a window into the future motion of the real robot. This information allows the user to avoid areas of the shared workspace that are soon to be inhabited by the robot. In addition, this information provides an early indication of the object (or group of objects) that will likely be the target. The temporal offset, or delay, between the virtual and real robots, is an adjustable parameter. These two visual signal modes allow four distinct mode combinations:

\begin{figure}
    \vspace{0.2cm}
    \centering
    \includegraphics[width=0.5\textwidth]{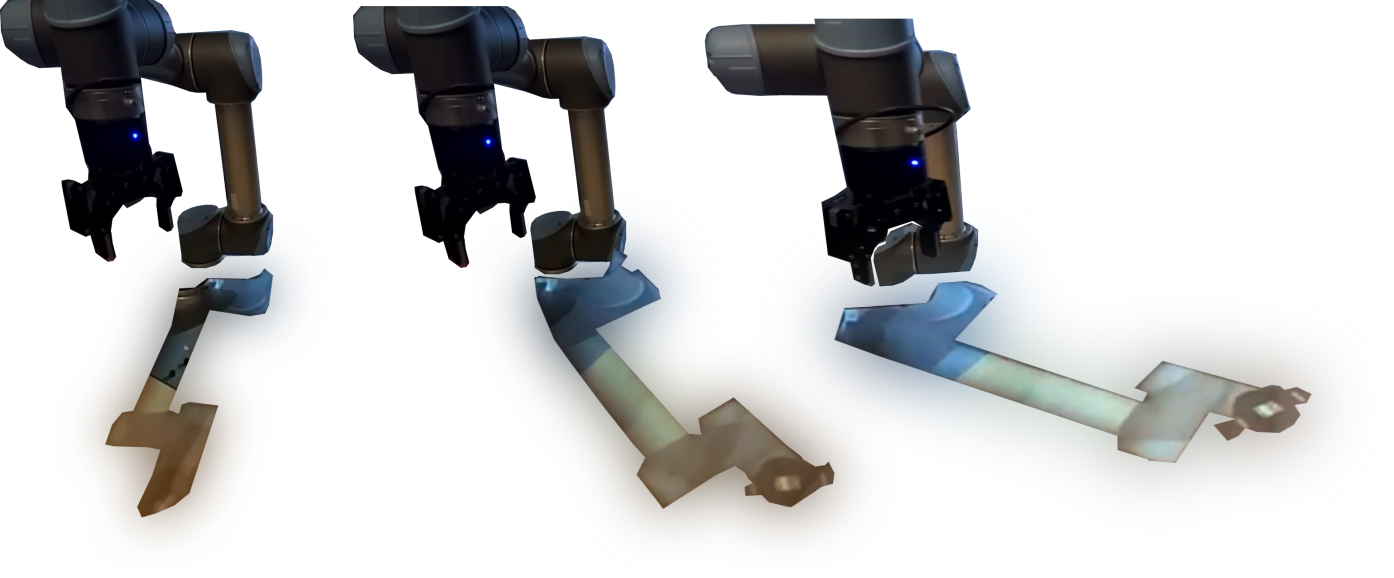}
    \caption{Dynamic signal: A virtual robot (bottom) moves ahead of the real robot resulting in a brief window for visualization of future motions.}
    \label{fig:shadow}
    \vspace{-0.5cm}
\end{figure}

\begin{enumerate}
    \item \textbf{No-Projection Mode}: This mode provides no visual cues and merely displays information about assigned cubes to the human subject on a laptop screen in the form of a grid with green colored squares. The six cubes assigned to the robot are shown in white.
    \item \textbf{Static Mode}: A visual cue showing a semi-circular disc is projected for one second to indicate the next target cube (out of six possible cubes). This visual information is intended to provide the human partner with information about the robot's next objective.
    \item \textbf{Dynamic Mode}:  A real-time rendered animation of a virtual robot is projected onto the table. This rendered digital twin previews the motion of the physical robot before they occur. A time delay of one second between the virtual and real robot arm is used. 
    \item \textbf{Dual Mode}: A hybrid mode combining visual cues from both Static and Dynamic Mode.  
\end{enumerate}

In our study, we utilized a 3 $\times$ 6 grid displayed on a screen to depict the location of 18 cubes on a table, as shown in Fig.~\ref{fig:teaser}. The grid comprises twelve green squares and six white squares that are randomly distributed across the grid. Users are tasked with picking up cubes located on the green squares, while the robot is assigned to collect cubes located on the white squares. The Static, Dynamic, or Dual Mode may provide users with visual cues about the robot's target objects.

\subsection{Measuring Information Transfer and Causality}
A critical question for identifying efficient signals for intention projection is how to measure their influence on human behavior. How much information transfer is there between the robot projecting its intent using a specific visual signal and the interacting human partner? How can this be quantified in an objective manner? 

To this end, we employ a formal, information-theoretic approach to describe information transfer. More specifically, we calculate the Transfer Entropy (TE)~\cite{schreiber2000measuring} between the sender (robot) and receiver (human). In this formulation, the visual signals form a communication channel between the two partners. TE measures the directed transfer of information between two processes and is widely used for inferring causal relationships between observed processes~\cite{staniek2008symbolic}. We can calculate it as:
$$
   \mathrm{TE}_{X \rightarrow Y} = H(Y_{t}|Y_{t-1:t-K}) - H(Y_{t}|Y_{t-1:t-K},X_{t-1:t-K})
$$
where $X$ and $Y$ are the source and target time series. In our specific case, $X$ corresponds to information about the robot projections (when did the robot project), whereas $Y$ corresponds to observations of the human behavior (when did the human pick an object). $H$ indicates the Shannon entropy while $K$ is a history length or past observations of the source time series. For best practices on how to set an optimal value of $K$ we refer the reader to \cite{bossomaier2016transfer}. In this case, we set the parameter to the average time window between two human pick events, i.e., $K=9$. For small sample sizes, the transfer entropy estimates are known to be biased~\cite{marschinski2002analysing}. To correct for bias, we use a specific variant of TE called Effective Transfer Entropy ($\widehat{\mathrm{TE}}$)~\ \cite{marschinski2002analysing}:
$$
    \widehat{\mathrm{TE}}_{X\rightarrow Y} = \mathrm{TE}_{X \rightarrow Y} - \frac{1}{\mathrm{Z}}\sum^{\mathrm{Z}}_{1} \mathrm{TE}_{X^{'}\rightarrow Y}
$$
where $\mathrm{Z} = 100$ as proposed in \cite{bossomaier2016transfer} and $X^{'}$ is the shuffled source time series. TE can be calculated in a data-driven fashion, i.e., by performing an experiment and collecting data about the timing of when certain visual signals were projected as well as data about when the human performed a certain action (e.g. a pick or lifting action).  

\section{Experiments and Results}
To compare the efficiency and impact of different visual cues, an  Institutional Review Board (IRB)\footnote{This study was approved by Arizona State University (\#STUDY00016445)}-approved human subject study was conducted with 22 subjects between the ages of 18-28. All participants voluntarily agreed to participate in the experiment, which was advertised as a sorting game with a robot in a flyer. They were not given any form of compensation for their participation. Additionally, participants were not provided with any information regarding the technical aspects or analysis of the experiment, except for the logistics of how the experiment would be conducted. Participants were asked to engage in the cube sorting task described in Section~\ref{subsec:task}. Throughout the task, the human-robot team has to sort all cubes simultaneously. Hence, the coordination of actions is critical for safety and efficiency. Since the study is conducted in a within-subjects (or repeated measures) manner, the order in which the four modes were introduced to each participant was randomized, making sure any potential bias or influence due to familiarity with the task can be minimized, allowing for a more accurate assessment of the effects of each mode on task execution.
In addition, the robot speed was set to variable safe speeds sampled from a Gaussian distribution with $\mu =0 .44$, $\sigma = 0.35$ (m/s). 

An overall underlying question in our experiments is whether the modes introduced above for visual signaling provide different and distinct degrees of improvement in the specific human-robot collaboration task discussed here. More specifically, the following hypotheses are investigated in this experiment:

\begin{table*}[t!]
\vspace{0.25cm}
\caption{Results on objective (Relative Finish Time) and subjective (NASA-TLX) measures using Wilcoxon signed rank test on pairs of proposed modes as their $p$ value. Here, the (+) and (-) notation results from whether the difference between the mean values of the compared modes (e.g. for Du vs St difference is calculated as $\mu$(Du)-$\mu$(St) for objective or subjective measures) is positive or negative. [Du = Dual, St = Static, Dy = Dynamic and No = No-Projection], [\sethlcolor{green}\hl{Statistically Significant},  \sethlcolor{white}\hl{Not Statistically Significant}]}
\label{Tab:wilcoxon_table}
\centering
\begin{tabular}{c|c||cccccc}
\hline
\hline
\multicolumn{1}{l|}{} & \cellcolor[HTML]{C0C0C0}\textbf{Objective Measure} ($p$ value)      & \multicolumn{6}{c}{\cellcolor[HTML]{C0C0C0}\textbf{Subjective Measures} ($p$ value)}                                                                                                                                                                         \\ \cline{1-8} 
                      & Relative Finish Time                   & Mental Demand                        & Physical Demand                    & Temporal Demand                   & Performance                        & Effort                             & Frustration                        \\ \hline
Du vs St              & \cellcolor[HTML]{FFFFFF}(+) 0.1907 & \sethlcolor{green}\hl{(-) 0.0062} & \cellcolor[HTML]{FFFFFF}(-) 0.4380  & \cellcolor[HTML]{FFFFFF}{(-) 0.0721} & \cellcolor[HTML]{FFFFFF}(-) 0.1154 & \cellcolor[HTML]{FFFFFF}(-) 0.2314 & \cellcolor[HTML]{FFFFFF}(-) 0.1157 \\
Du vs Dy              & \cellcolor[HTML]{FFFFFF}(+) 0.2760  & \cellcolor[HTML]{FFFFFF}(-) 0.6163 & \cellcolor[HTML]{FFFFFF}(-) 0.4669 & \cellcolor[HTML]{FFFFFF}(-) 0.1441 & \cellcolor[HTML]{FFFFFF}(-) 0.3740  & \cellcolor[HTML]{FFFFFF}(-) 0.3232 & \cellcolor[HTML]{FFFFFF}(-) 0.2467 \\
Du vs No              & \sethlcolor{green}\hl{(-) 0.0101}   & \sethlcolor{green}\hl{(-) 0.0007} & \cellcolor[HTML]{FFFFFF}(-) 0.1350  & \sethlcolor{green}\hl{(-) 0.0039} & \cellcolor[HTML]{FFFFFF}(-) 0.1081 &  \sethlcolor{green}\hl{(-) 0.0135} & \sethlcolor{green}\hl{(-) 0.0191} \\
St vs Dy              & \cellcolor[HTML]{FFFFFF}(+) 0.4120  & \cellcolor[HTML]{FFFFFF}(+) 0.3591 & \cellcolor[HTML]{FFFFFF}(+) 0.4577 & \cellcolor[HTML]{FFFFFF}(+) 0.5701   & \cellcolor[HTML]{FFFFFF}(+) 0.8885 & \cellcolor[HTML]{FFFFFF}(+) 0.4308 & \cellcolor[HTML]{FFFFFF}(+) 0.6182 \\
St vs No              &  \sethlcolor{green}\hl{(-) 0.0481} & \cellcolor[HTML]{FFFFFF}{(-) 0.0551} & \cellcolor[HTML]{FFFFFF}(-) 0.3104 & \cellcolor[HTML]{FFFFFF}(-) 0.3464 & \cellcolor[HTML]{FFFFFF}(-) 0.3130  & \cellcolor[HTML]{FFFFFF}(-) 0.3307 & \cellcolor[HTML]{FFFFFF}{(-) 0.0770}  \\
Dy vs No              & \cellcolor[HTML]{FFFFFF}(+) 0.2029 & \cellcolor[HTML]{FFFFFF}{(-) 0.0754} & \cellcolor[HTML]{FFFFFF}(-) 0.3217 & \cellcolor[HTML]{FFFFFF}(-) 0.2044 & \cellcolor[HTML]{FFFFFF}(-) 0.9629 &  \sethlcolor{green}\hl{(-) 0.0104} &  \sethlcolor{green}\hl{(-) 0.0254} \\ \hline \hline
\end{tabular}
\end{table*}

\begin{itemize}
    \item \textbf{H1}: At least one projection mode enhances task efficiency compared to the No-projection mode.    
    \item \textbf{H2}: Cognitive load indices are substantially lower in projection modes compared to the No-projection mode.
\end{itemize}

To provide evidence for or against the above hypotheses, we combined both subjective and objective analyses. Participants were asked to fill in the NASA-TLX (Task Load Index) questionnaire~\cite{feick2020virtual} after experimenting with each mode, which comprises six sub-scales (\textit{Mental Demand, Physical Demand, Temporal Demand, Performance, Effort,} and \textit{Frustration}) rated on a 0-20-point scale. However, for visualization purposes these scores were normalized between 0-100. To evaluate data distribution, subjective and objective measures were analyzed with the Friedman test, as described in \cite{de2021leveraging}. In particular, the Friedman test was used to determine if there were any significant differences in the distribution of data across the four modes. Subsequently, a Wilcoxon signed rank test, as described in \cite{hirschmanner2019virtual}, was conducted to identify any statistically significant differences between the four modes. The estimated $p$ values were adjusted with a Bonferroni correction to prevent Type-I error. These analyses are discussed in detail respectively in Sec.~\ref{sec:temporal} and Sec.~\ref{sec:subjective}.

\subsection{Hypothesis 1: Task Efficiency}
\label{sec:temporal}
\begin{figure}[h!]
    \centering
    \vspace{-0.5cm}
    \includegraphics[width= 0.5\textwidth]{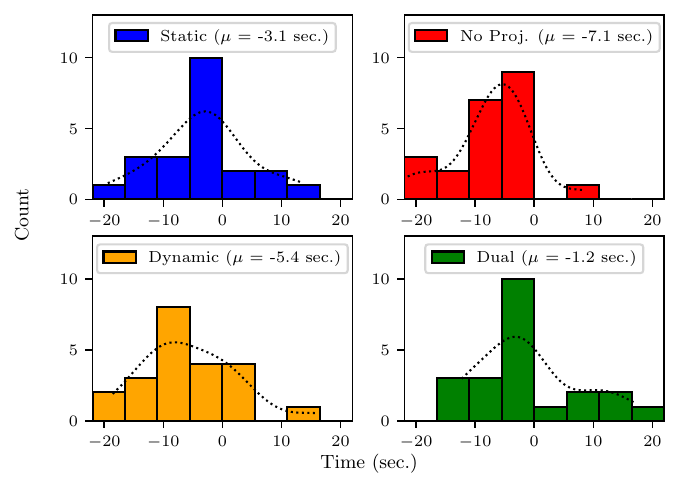}
    \caption{Histogram of $\Delta t$ between robot vs human finish times. A positive $\Delta t$ indicates that the subject finished earlier than the robot. In other words, the efficiency was higher. From this figure, there is an explicit lean toward the right in Dual mode, indicating a higher efficiency than other modes.}
    \label{fig:rel_hfinish_time}
\end{figure}

To investigate task efficiency, we analyze the relative finish time of the robot with respect to the human interaction partner across all users and modes. Relative finish time was used to gauge human-robot collaboration effectiveness. This decision was influenced by its capability to account for varying task conditions like robot speed fluctuations and visual signal changes. It allowed to quantitatively determine which visual signals significantly enhanced human task efficiency in the collaborative scenario. We calculate the relative finish time  $\Delta t = \left(t_{r} - t_{h}\right)$ where $t_{r}$ is the robot finish time, and $t_{h}$ is the human finish time. A positive value for $\Delta t$ indicates that the participant finished the sorting task before the robot. A negative value of $\Delta t$ indicates that the human was slower than the robot at sorting. Fig.~\ref{fig:rel_hfinish_time} shows the distributions of relative finish times across all modes, and after passing the Friedman test ($p$ < 0.05) over all four modes, the corresponding $p$ values of paired Wilcoxon signed ranked tests with regards to $\Delta t$ were calculated and shown in Table~\ref{Tab:wilcoxon_table}. Based on the results presented in Fig. \ref{fig:rel_hfinish_time}, it can be observed that the mean $\Delta t$ is higher for the Dual mode ($-1.2$~sec.) when compared to the No-projection mode ( $-7.1$~sec.). This result has been statistically validated in the significance test presented in Table~\ref{Tab:wilcoxon_table} ($p$ < 0.05). Conversely, the difference in $\Delta t$ between the Static and Dynamic modes was found to be insignificant ($p$ > 0.05), despite the mean value of $\Delta t$ being higher in the Static mode than in the Dynamic mode. Additionally, the Static mode exhibited a significantly faster $\Delta t$ on average (56~\%) compared to the No-projection mode ($p$ < 0.05). 

These findings support hypothesis \textbf{H1}: both the Static and Dual modes show a significant difference when compared to the No-projection mode. Significant improvements in task efficiency can be observed -- especially in the case of the Dual mode. Compared to that, the Dynamic mode shows no significant difference from the No-projection mode.

\subsection{Hypothesis 2: Cognitive Load}
\begin{figure}[h]
\vspace{-0.2cm}
    \centering
    \includegraphics[width=0.5\textwidth]{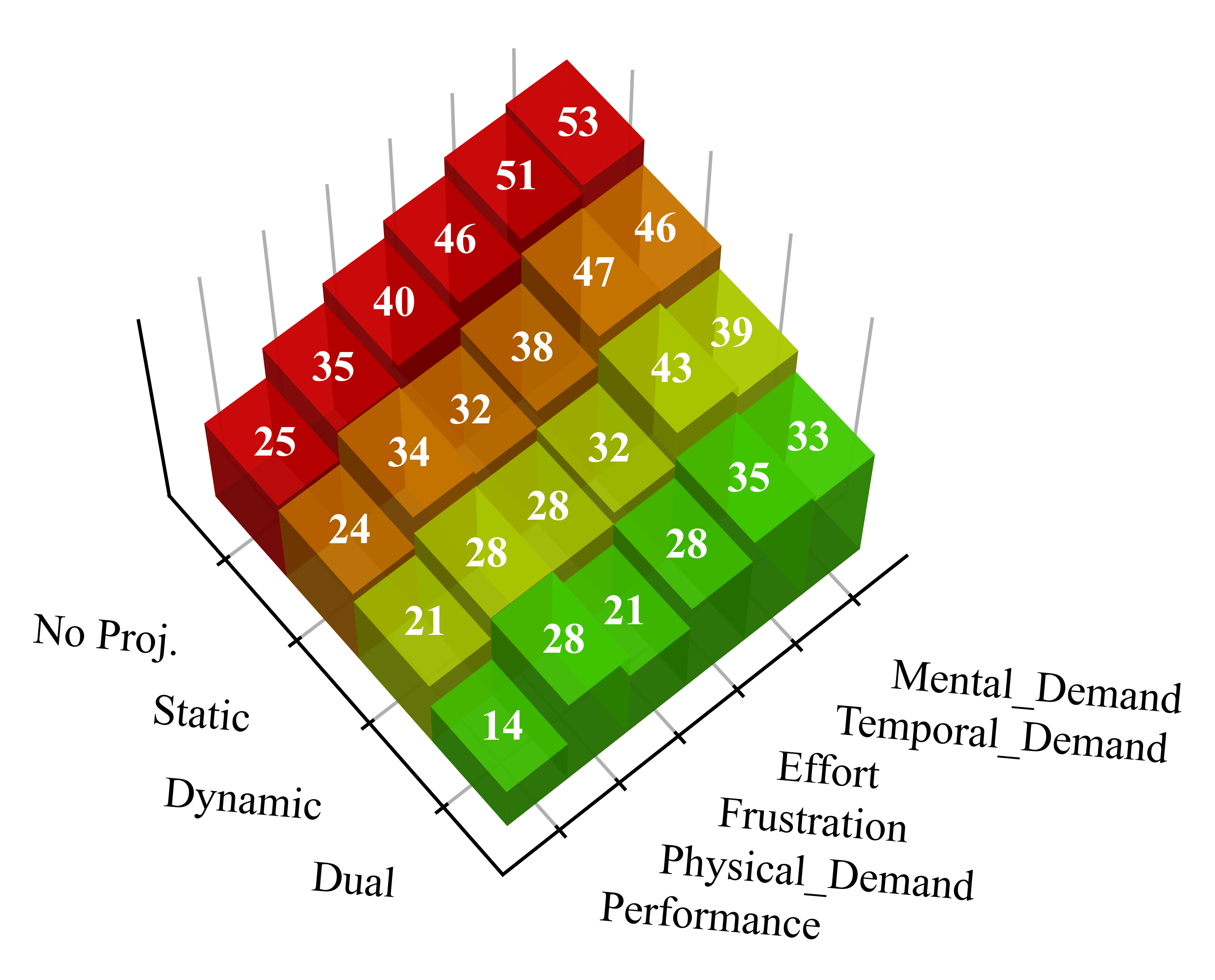}
    \caption{Visualization of all subjective mental workload indices across all the modes.}
    \label{fig:nasa_tlx}
\vspace{-0.5cm}
\end{figure}

\vspace{-0.1cm}
\label{sec:subjective}
To address the second hypothesis, we focus on the subjective feedback provided by users in the form of NASA TLX scores. Fig.~\ref{fig:nasa_tlx} shows a summary of mean scores across all modes and workload indices (lower values correspond to better subjective responses). Looking at Fig.~\ref{fig:nasa_tlx}, we notice that all projection modes consistently produce better scores when compared to the No-projection mode. The Dual mode (combining visual cues) substantially decreases all indices, e.g., frustration is reduced by 48\%, the mental demand is reduced by 37\% and effort sees a 39\% reduction compared to No-projection mode. Furthermore, all cognitive load indices across all four modes show significant differences in the subjective score as verified by the Friedman test ($p$ < 0.05). Table~\ref{Tab:wilcoxon_table} provides additional details regarding the pairwise significance of the four modes across six different cognitive load indices. With regards to \textit{Mental Demand}, we find that the Dual mode results in significantly better scores (highlighted in green) when compared to the Static projection and No-projection modes ($p$ < 0.05). However, when looking at the difference between the mean values of compared modes, we note that both the Dual and Dynamic modes performed better than the Static and No-projection modes, as indicated in Table~\ref{Tab:wilcoxon_table}. Interestingly, the Dynamic mode showed better scores than the Static mode. Contrary to what we observed for relative finish times in Sec.~\ref{sec:temporal}, the Dynamic mode seems to provide marginal improvements in reducing cognitive load.


On the other hand, cognitive load indices such as \textit{Physical Demand} and \textit{Performance} did not see any statistically significant change in scores ($p$ > 0.05). However, given that the sorting task remained the same across all four modes, with the only variation being the information feedback in the form of visual signals to human subjects, it is reasonable to assume that their \textit{Performance} and \textit{Physical Demand} would remain similar across all four modes. 
Regarding \textit{Effort} and \textit{Frustration}, we find that both the Dynamic and Dual modes have statistically significant scores (highlighted in green) when compared to the No-projection mode ($p$ < 0.05). Furthermore, although the Static mode did not demonstrate a significant improvement in terms of frustration and effort when compared to the No-projection mode, the differences in their mean scores indicate that human subjects still preferred the Static mode over the No-projection mode. Finally, \textit{Temporal Demand} shows that Dual mode has a statistically significant (lower) score when compared to No-projection mode, which means that human subjects were not hurried or rushed by the robot's actions and were able to finish tasks with ease. Nonetheless, the Dual mode is again slightly better than the Static mode, i.e., differences in mean and close to significant $p$ value  in the first row of \textit{Temporal Demand} column in Table \ref{Tab:wilcoxon_table}.

In summary, our hypothesis \textbf{H2} is partially supported by the evidence: both the Dual and Dynamic modes see significant or close to significant reductions in load indices with respect to \textit{Mental Demand, Temporal Load, Effort}, and \textit{Frustration}. With regard to these cognitive load indices, the Dynamic mode marginally outperforms the Static mode.

\subsection{Transfer Entropy Analysis}
The insights and conclusions drawn above are based on objective metrics (relative finish time) and subjective human feedback (cognitive load indices). In this section, we investigate if similar insights can be drawn without any insight into the task, i.e., purely from the motion data of both humans and the robot. 

More specifically, we use a (task-agnostic) information-theoretic metric to analyze recorded data during experiments and evaluate whether the findings corroborate the results observed in Sec.~\ref{sec:temporal} and Sec.~\ref{sec:subjective}. We conducted a posthoc analysis using Transfer Entropy~\cite{schreiber2000measuring, marschinski2002analysing} and videos collected during the experiment. We manually annotated time stamps in which one of the following events/actions occur: human pick ($\mathrm{hp}$), robot stop ($\mathrm{rs}$), virtual robot stop ($\mathrm{vs}$), and Static goal ($\mathrm{sg}$). Events ($\mathrm{hp}$) indicate timestamps at which the human picked a cube. Similarly, ($\mathrm{rs}$) indicates timestamps wherein the real robot stops moving to pick up an object. Events ($\mathrm{vs}$) indicate timestamps in which the \emph{virtual} robot stopped moving, while the event ($\mathrm{sg}$) shows the time stamp at which the Static goal was projected. These four actions result in four separate time series $\mathrm{\tau}_{\text{a}}$ with binary values of 0 or 1, depending on whether the respective action occurred or not. For Static and No-projection mode, ($\mathrm{vs}$) will be a time series of zero values. We define each participant as $\mathrm{P}_{n}$ where $n \in \{1,2,...., N\}$. For each participant, we define time series for four actions ($a$) per mode:
$$
\mathrm{\tau}_a(m) = [v_{1},v_{2},...,v_{T}]\text{ where } a \in \{\mathrm{hp},\mathrm{rs},\mathrm{vs}, \mathrm{sg}\},
$$
with $m \in \{\text{No-Projection},\text{Dynamic},\text{Static},\text{Dual}\}$ and $T$ being the number of time steps. The variables $v_1, v_2, ..., v_t$ are binary variables indicating whether the action $a$ occurs at the current timestamp ($t$). 

\begin{table}[th]
\vspace{0.15cm}
\caption{Average Effective Transfer Entropy Across Different Modes ($m$) for annotated time series.}
\label{Tab:ETE}
\begin{center}
\scalebox{0.95}{
\begin{tabular}{l c c c c }
    \toprule
    &No Proj.
    &Static
    &Dynamic
    &Dual
    \\
    \midrule
    $\widehat{\mathrm{TE}}_{[\mathrm{vs}]\rightarrow [\mathrm{hp}]}$
    & 0.0   & 0.0   & \textbf{0.00864755}  & \textbf{0.0148222} 
    \\
    $\widehat{\mathrm{TE}}_{[\mathrm{sg}]\rightarrow [\mathrm{hp}]}$
    & 0.0    & \textbf{0.01296397}   & 0.0  &  0.00418501
    \\
     $\widehat{\mathrm{TE}}_{[\mathrm{rs}]\rightarrow [\mathrm{hp}]}$
    & 0.00968205    & 0.00919081   & 0.00484169   &  0.01176273
    \\
\bottomrule
\end{tabular}}
\end{center}
\vspace{-0.35cm}
\end{table}

Based on the above time series we compute the effective Transfer Entropy for different modes as seen in Table~\ref{Tab:ETE}. Here, $\widehat{\mathrm{TE}}_{[\mathrm{vs}]\rightarrow [\mathrm{hp}]}$ signifies the information transfer between the virtual robot stopping and the human picking an object, since this visual cue is only used in the Dynamic and Dual modes, we only observe an information transfer in either one of these modes. $\widehat{\mathrm{TE}}_{[\mathrm{sg}]\rightarrow [\mathrm{hp}]}$ shows information transfer between the static goal (signal) and human picking action and final $\widehat{\mathrm{TE}}_{[\mathrm{rs}]\rightarrow [\mathrm{hp}]}$ shows information transfer between physical robot and human picking. The $\widehat{\mathrm{TE}}$\footnote{General notation for Effective Transfer Entropy. Includes all the different combinations of source and target time series show in Table~\ref{Tab:ETE}.} of about $0.0086$ for the virtual robot in the Dynamic mode is higher than the real robot in the No-projection mode with $\widehat{\mathrm{TE}}= 0.0091$. Moreover, an even higher value of $0.0129$ can be achieved when projecting a static signal. The highest overall value for the effective Transfer Entropy of $0.0148$ is observed in the Dual mode -- which again emphasizes the power of projecting multiple visual cues in conjunction. 

The $\widehat{\mathrm{TE}}$ between the real robot and human picks never exceeds $0.011$, i.e., robot actions influence human behavior to a lesser degree than the visual signals. Generally, it can be noticed that the information transfer is highest between visual signals and human actions. The overall trend identified via Transfer Entropy mirrors the similar trends found in the earlier analysis of relative finish times and subjective mental workload assessment: the Dual mode shows the best performance, followed by the individual projection modes (Static and Dynamic), and finally, the No-projection mode.

\section{Discussion and Limitations}
Contrary to previous assumptions, this paper found that projecting combination of visual cues during human-robot interactions significantly improves collaborative task performance. Objective, subjective, and information-theoretic metrics all support this conclusion. Moreover, carefully designing static and dynamic visual cues can enhance collaborative tasks. Nevertheless, further research is necessary to explore the boundaries of this finding. For instance, it remains to be seen whether there is an upper bound to the effective design and combination of these visual cues that, if exceeded, might hamper collaborative task efficiency. One potential avenue for future research is to investigate whether the visual signals convey different aspects of the task. For the current study, a possible rationalization about the design aspect of signals is that the static signal communicates the robot's intended destination. In contrast, the dynamic signal conveys how the robot will reach that location. Additionally, while the time delay between virtual and real robot trajectories was constant in this study, future investigations could explore the impact of variable time delays on human-robot collaboration.

The results of our experiment showed a clear distinction between Dual and No-projection mode. However, there was no significant distinction between dynamic and static signals. We suggest conducting a more careful investigation of the experiment dimension to better understand the extent of the role of visual signals in human-robot interactions. Specifically, further research should focus on studying the type and appearance of visual signals to confirm their vital role in affecting how human subjects interact with the robot system. 

Finally, our investigation on Transfer Entropy indicated that information-theoretic measures are able to provide early indications regarding the amount of information transferred by a visual cue to human users. Nevertheless, more in-depth studies are required to confirm the effectiveness of Transfer Entropy on a broader setup in the future.

\bibliographystyle{IEEEtran}
\scriptsize{\bibliography{references}}

\end{document}